\documentclass[pdflatex,sn-mathphys-num,iicol]{sn-jnl}

\usepackage[T1]{fontenc}
\usepackage{graphicx}
\usepackage{multirow}
\usepackage{amsmath,amssymb,amsfonts}
\usepackage{xcolor}
\usepackage{booktabs}
\usepackage{tabularx}
\usepackage{array}
\usepackage[most]{tcolorbox}
\usepackage[title]{appendix}
\usepackage{hyperref}
\usepackage{tikz}
\usetikzlibrary{arrows.meta,positioning,fit,calc}
\newcommand{\fraudattempt}{\emph{fraudulent attempt}}


\newcommand{\BestFSMF}{0.852}
\newcommand{\BestFSName}{Gemma 4 31B (ASU)}

\newcommand{\CVBestLLMMF}{0.820}
\newcommand{\CVBestLLMName}{Claude 5 Sonnet}

\newcommand{\CVJointMF}{0.857}
\newcommand{\CVJointSD}{0.038}

\newcommand{\CallsSaved}{2\%}
\newcommand{\ClaudeGenuine}{50\%}
\newcommand{\ClaudeRecall}{55\%}
\newcommand{\CompBreakMax}{54\%}
\newcommand{\CompBreakMin}{2\%}
\newcommand{\CompEnsAcc}{0.875}
\newcommand{\CompFN}{23}
\newcommand{\CompFNRightMax}{78\%}
\newcommand{\CompFNRightMin}{48\%}
\newcommand{\CompFP}{27}
\newcommand{\CompFixMax}{59\%}
\newcommand{\CompFixMin}{11\%}
\newcommand{\CompOracleMax}{0.953}
\newcommand{\CompOracleMin}{0.930}
\newcommand{\CompTP}{192}
\newcommand{\DBEmb}{0.134}
\newcommand{\DBEmbHi}{0.202}
\newcommand{\DBEmbLo}{0.064}

\newcommand{\DBRule}{0.274}
\newcommand{\DBRuleHi}{0.329}
\newcommand{\DBRuleLo}{0.217}
\newcommand{\DBRuleP}{$=$0.002}
\newcommand{\DBWin}{0.272}

\newcommand{\DBWinP}{$=$0.002}
\newcommand{\DLateRule}{32\%}
\newcommand{\DLateRuleHi}{47\%}
\newcommand{\DLateRuleLo}{18\%}
\newcommand{\DLateRuleP}{$\le$0.001}
\newcommand{\DLateWin}{24\%}
\newcommand{\DLateWinHi}{42\%}
\newcommand{\DLateWinLo}{8\%}
\newcommand{\DLateWinP}{$=$0.016}
\newcommand{\DeltaDualMax}{0.095}
\newcommand{\DeltaDualMaxP}{0.021}
\newcommand{\DeltaDualMin}{0.043}
\newcommand{\DeltaNoOcddMax}{+0.005}
\newcommand{\DeltaNoOcddMin}{-0.008}
\newcommand{\EWGainMax}{0.010}
\newcommand{\EWLossMax}{0.068}

\newcommand{\EWMsgs}{6.8}

\newcommand{\ExtCompFN}{18}
\newcommand{\ExtCompFNRightMax}{83\%}
\newcommand{\ExtCompFNRightMin}{44\%}
\newcommand{\ExtDeltaDualMax}{0.185}
\newcommand{\ExtDeltaDualMin}{0.130}
\newcommand{\ExtDeltaNoOcddMax}{-0.003}
\newcommand{\ExtDeltaNoOcddMin}{-0.004}

\newcommand{\ExtDriftShare}{99\%}
\newcommand{\ExtIndSVM}{0.820}

\newcommand{\ExtJointHi}{0.908}
\newcommand{\ExtJointLo}{0.868}
\newcommand{\ExtJointMF}{0.888}
\newcommand{\ExtJointPrec}{0.838}
\newcommand{\ExtJointRec}{0.964}
\newcommand{\ExtLLMMax}{0.754}
\newcommand{\ExtLLMMin}{0.685}
\newcommand{\ExtLLMPrecMax}{0.750}
\newcommand{\ExtLLMPrecMin}{0.636}
\newcommand{\ExtLLMRecMax}{0.988}
\newcommand{\ExtLLMRecMin}{0.690}

\newcommand{\ExtPTwoDualMax}{0.755}
\newcommand{\ExtPTwoDualMin}{0.675}
\newcommand{\ExtPTwoJointMax}{0.885}
\newcommand{\ExtPTwoJointMin}{0.853}
\newcommand{\ExtPTwoMaxP}{$<$0.001}
\newcommand{\FactNormEmb}{0.715}
\newcommand{\FactNormTf}{0.581}
\newcommand{\FactNormTfHi}{0.635}
\newcommand{\FactNormTfLo}{0.522}
\newcommand{\FactWinEmb}{0.439}
\newcommand{\FactWinTf}{0.443}
\newcommand{\FewShotK}{4}

\newcommand{\GemmaDualCalls}{1.58}
\newcommand{\GemmaDualCost}{0.31}

\newcommand{\GemmaJointCalls}{0.54}
\newcommand{\GemmaJointCost}{0.13}

\newcommand{\JointAcc}{0.808}
\newcommand{\JointDriftN}{214}
\newcommand{\JointDriftShare}{98\%}
\newcommand{\JointFlagged}{219}

\newcommand{\JointIndMF}{0.554}
\newcommand{\JointMF}{0.806}
\newcommand{\JointMFhi}{0.846}
\newcommand{\JointMFlo}{0.762}
\newcommand{\JointMembers}{SVM, Random Forest and kNN}

\newcommand{\LCDiffForty}{0.139}

\newcommand{\LCDiffThreeTwenty}{0.049}
\newcommand{\LCIndEighty}{0.729}
\newcommand{\LCIndForty}{0.602}
\newcommand{\LCIndOneSixty}{0.752}
\newcommand{\LCIndThreeTwenty}{0.825}

\newcommand{\LCTransEighty}{0.813}
\newcommand{\LCTransForty}{0.742}
\newcommand{\LCTransOneSixty}{0.838}
\newcommand{\LCTransThreeTwenty}{0.874}
\newcommand{\LLMZSMax}{0.828}
\newcommand{\LLMZSMin}{0.747}
\newcommand{\LateN}{38}
\newcommand{\LateOcdd}{32\%}
\newcommand{\LateOcddHi}{47\%}
\newcommand{\LateOcddLo}{18\%}
\newcommand{\LateRule}{0\%}
\newcommand{\LateWin}{8\%}

\newcommand{\LocWorse}{1}
\newcommand{\MaxMsgs}{20}
\newcommand{\MedianMsgs}{11}
\newcommand{\MedianOnsetFraud}{2}
\newcommand{\MedianOnsetGenuine}{4}
\newcommand{\MinMsgs}{3}
\newcommand{\NAmbHigh}{57}
\newcommand{\NBen}{185}

\newcommand{\NConv}{400}
\newcommand{\NDriftAll}{316}
\newcommand{\NDriftFraud}{212}
\newcommand{\NDriftGenuine}{104}
\newcommand{\NDriftNone}{84}
\newcommand{\NExtJudges}{5}
\newcommand{\NExtLLMsig}{10}
\newcommand{\NExternal}{1000}
\newcommand{\NJointSig}{9}

\newcommand{\NJudgeExt}{5}
\newcommand{\NJudgeInput}{5}
\newcommand{\NJudges}{9}
\newcommand{\NLLM}{10}

\newcommand{\NMal}{215}
\newcommand{\NNoDriftAll}{84}
\newcommand{\NScenCollab}{160}
\newcommand{\NScenOther}{80}
\newcommand{\NTest}{360}
\newcommand{\NTestBen}{169}
\newcommand{\NTestMal}{191}
\newcommand{\NTrain}{40}
\newcommand{\NTrainBen}{16}
\newcommand{\NTrainMal}{24}

\newcommand{\NovaBlocked}{36\%}
\newcommand{\OcddBal}{0.715}
\newcommand{\OcddBalHi}{0.770}
\newcommand{\OcddBalLo}{0.656}
\newcommand{\OcddDelay}{2}
\newcommand{\OcddDetect}{93\%}
\newcommand{\OcddDims}{10}
\newcommand{\OcddFalse}{35\%}
\newcommand{\OcddFoldsSame}{5}
\newcommand{\OcddLocal}{78\%}

\newcommand{\OcddOffBal}{0.486}
\newcommand{\OcddOffBalHi}{0.543}
\newcommand{\OcddOffBalLo}{0.428}

\newcommand{\OcddQRange}{0.8--0.9}

\newcommand{\OffMinP}{0.480}
\newcommand{\OffNJointWins}{7}
\newcommand{\OldBal}{0.443}
\newcommand{\OldFalse}{99\%}
\newcommand{\OldLocal}{87\%}
\newcommand{\PTwoBestCost}{0.13}
\newcommand{\PTwoBestHi}{0.904}
\newcommand{\PTwoBestJudge}{Gemma 4 31B (ASU)}
\newcommand{\PTwoBestLo}{0.839}
\newcommand{\PTwoBestMF}{0.872}

\newcommand{\PTwoPhaseOne}{0.874}

\newcommand{\RouteCVMax}{+0.005}
\newcommand{\RouteCVMin}{-0.017}

\newcommand{\RouteExtMax}{-0.013}
\newcommand{\RouteExtMin}{-0.023}
\newcommand{\RouteExtN}{5}
\newcommand{\RouteExtNSig}{3}
\newcommand{\RouteOffMax}{+0.043}
\newcommand{\RouteOffMin}{+0.024}
\newcommand{\RouteOffMinP}{0.120}
\newcommand{\RouteShare}{14\%}
\newcommand{\RuleBal}{0.441}

\newcommand{\ShareFraudOnsetTwo}{69\%}
\newcommand{\ShareOnsetTwoFour}{87\%}
\newcommand{\SklearnVersion}{1.9.1}
\newcommand{\TypeFraudMax}{99\%}
\newcommand{\TypeFraudMin}{62\%}
\newcommand{\TypeGenMax}{50\%}
\newcommand{\TypeGenMin}{0\%}
\newcommand{\WordsPerMsg}{47}

\begin{document}

\title[Joint Detection of Fraud and Concept Drift in Online Conversations]{Joint Detection of Fraud and Concept Drift in Online Conversations with LLM-Assisted Judgment}

\author*[1,2]{\fnm{Ali} \sur{{\c{S}}enol}}\email{alisenol@tarsus.edu.tr; asenol@asu.edu}
\author[2,3]{\fnm{Garima} \sur{Agrawal}}\email{garima.agrawal@asu.edu}
\author[2]{\fnm{Huan} \sur{Liu}}\email{huanliu@asu.edu}

\affil*[1]{\orgdiv{Department of Computer Engineering}, \orgname{Tarsus University}, \orgaddress{\city{Tarsus}, \state{Mersin}, \postcode{33400}, \country{T\"urkiye}}}
\affil[2]{\orgdiv{School of Computing and Augmented Intelligence}, \orgname{Arizona State University}, \orgaddress{\city{Tempe}, \state{AZ}, \postcode{85281}, \country{USA}}}
\affil[3]{\orgname{Minerva CQ and HumaConn AI Consulting}, \orgaddress{\city{San Tan Valley}, \state{AZ}, \postcode{85140}, \country{USA}}}

\abstract{Social-engineering conversations often open as plausible professional outreach and only later request personal or financial information. We study a two-phase framework: a majority-vote ensemble of classical models flags suspicious conversations, a One-Class Drift Detector (OCDD) locates the shift within each flagged dialogue, and a large language model (LLM) judges whether it is a fraudulent attempt or a genuine change. We compare this Joint framework with a Dual-LLM framework, in which one LLM performs both phases, using \NLLM{} current LLMs, the human-annotated SEConvo corpus and held-out conversations. An ensemble trained on few labels is as accurate as the strongest LLM classifiers and generalizes better to held-out conversations from the same corpus, where LLMs flag many legitimate exchanges. Confining LLM calls to flagged conversations makes the Joint framework more accurate and cheaper than Dual-LLM, but the gain comes from the filter: the judge adds a drift-type label rather than accuracy, since it never sees the attacks the ensemble misses, many of which LLMs recognize as classifiers. In short dialogues, the reference a message is compared with decides whether drift is detectable: normal messages pooled across training dialogues localize the annotated shift, whereas the dialogue's own history is no better than a fixed-position rule. In this corpus, where requests come early, the detector's output leaves the final decision unchanged, and LLM judges tend to read legitimate requests as fraud. LLMs are thus most useful where they add information a supervised filter lacks, not as universal detectors or second checks of its alarms.}


\maketitle

\section{Introduction}\label{sec:intro}

Professional networking platforms are also channels for social engineering. A typical attack opens with a credible, often flattering message about a collaboration, a grant, an interview or a job, and only gradually steers the exchange toward the victim's date of birth, address, identity documents or payment details~\cite{ai2024defending}. Unlike fake-review detection~\cite{he2024detecting,alshehri2024online} or fraud detection in telephone and identity-verification dialogues~\cite{bajaj2019fraud,wang2019you}, the unit of analysis is a multi-turn dialogue whose intent becomes visible only as it unfolds, and a change of topic is not in itself evidence of fraud: legitimate conversations also move to logistics and may contain proportionate requests for contact details.

Existing approaches leave three gaps. Text classifiers that score a whole conversation can be accurate, but they do not show where the conversation turns or whether the turn is malicious. LLMs can read and explain a dialogue, yet using them on every conversation is costly, exposes sensitive content to external services, and, as we show, makes them flag many legitimate exchanges. Concept-drift detectors could locate the turn, but they were designed for long data streams and are rarely tested inside dialogues of about ten messages or against an annotated onset, so it is unknown whether they detect anything beyond the position at which such requests usually occur.

We address these gaps with a framework that assigns each task to the component suited to it. A lightweight supervised ensemble flags suspicious conversations, so that LLMs are called only for a selected subset of conversations. Within each flagged dialogue, a One-Class Drift Detector (OCDD)~\cite{gozuaccik2021concept} built on a reference of normal messages locates the shift, and an LLM judges whether the shift is a fraudulent attempt or a genuine change. We compare this \emph{Joint} framework with a \emph{Dual-LLM} framework in which an LLM replaces the ensemble and also acts as the judge. Every component is evaluated separately against the human annotation, including the message at which personal information is first requested, with trivial baselines, cross-validation and held-out data, so that the contribution of each part can be measured rather than assumed.

Formally, a conversation $C=(m_1,\dots,m_T)$ is a sequence of messages between an initiator and a recipient. Given a small set of labelled conversations, the task is to decide whether $C$ is malicious, to locate the message $t_0$ at which a suspicious conversation shifts, and to decide whether that shift is fraudulent or genuine, while keeping the use of LLMs selective. We address three research questions:
\begin{itemize}
    \item \textbf{RQ1.} How accurately does the ensemble flag malicious conversations compared with current LLM classifiers, on the annotated data and on unseen conversations?\\
    \emph{H1:} trained on few labelled conversations, the ensemble is at least as accurate as zero- and few-shot LLMs and generalizes better to unseen conversations.
    \item \textbf{RQ2.} Can OCDD localize the shift inside a flagged dialogue, and can an LLM judge tell fraudulent from genuine shifts?\\
    \emph{H2:} OCDD fires at or shortly after the annotated onset and rarely in dialogues without a shift, and the judge recovers the annotated drift type.
    \item \textbf{RQ3.} How do the Joint and Dual-LLM frameworks compare in final quality and cost?\\
    \emph{H3:} the Joint framework is at least as accurate as Dual-LLM at a lower LLM cost.
\end{itemize}

\noindent The main contributions of this work are:
\begin{itemize}
    \item a Joint framework that combines a supervised filter, within-dialogue drift detection and selective LLM judgment, with every model choice tuned on training labels only and the text representations fitted without labels;
    \item a drift detector for short dialogues, OCDD with a reference of normal messages and sentence embeddings, which detects and localizes the shift well above a fixed-position baseline, whereas a reference built from the dialogue's own history does not;
    \item a comparison with \NLLM{} current LLMs as classifiers and judges, with cross-validation, paired statistical tests, cost accounting and a held-out evaluation on \NExternal{} additional conversations, showing that the Joint framework is significantly more accurate and cheaper than Dual-LLM;
    \item controlled ablations that show where each component helps: the gain over Dual-LLM comes from the supervised filter, the final decision is no more accurate than that of the ensemble alone, the detector's output does not change the judge's decision in this corpus, and LLM judges tend to read legitimate requests as fraud;
    \item an error and learning-curve analysis showing that the LLMs recognize many attacks the ensemble misses, which a simple routing rule exploits only with few labels, and that the transductive advantage shrinks as labels grow.
\end{itemize}

\noindent The rest of the paper is organized as follows. Section~\ref{sec:related} reviews work on deception and fraud detection, concept drift detection and LLMs as classifiers and judges. Section~\ref{sec:method} describes the Joint and Dual-LLM frameworks, and Section~\ref{sec:setup} the data, evaluation protocols and statistical analysis. Section~\ref{sec:results} answers the three research questions, Section~\ref{sec:discussion} discusses the findings and limitations, and Section~\ref{sec:conclusion} concludes the paper.

\section{Related Work}\label{sec:related}

\paragraph{Deception and fraud detection.}
Opinion-spam detection moved from lexical and behavioural cues~\cite{jindal2008opinion,mukherjee2012spotting} to transformer-based classifiers~\cite{mohawesh2024fake}. In dialogue, linguistic features detect fraud in financial-service calls~\cite{bajaj2019fraud}, interactive questioning exposes identity fraud~\cite{wang2019you}, and content cues predict fraud on crowdfunding platforms~\cite{siering2016detecting}. Such text classifiers assume that a static representation of the whole text suffices~\cite{mironczuk2023twenty} and do not explain where a conversation changes~\cite{zhang2022adaptive}. SEConvo~\cite{ai2024defending,seconvo2024data} was built for social-engineering dialogues; its authors evaluate zero- and few-shot LLMs and propose ConvoSentinel, which analyses individual messages and retrieves similar annotated conversations. We use their split and metric, so their results serve as an external reference point (Section~\ref{sec:res-phase1}). {\c{S}}enol et al.~\cite{senol2026domain} enrich LLM prompts with domain knowledge distilled from the fraud-detection literature and combine such LLMs with OCDD in a pipeline that flags deceptive conversations, monitors them for drift and classifies the drift type. In that design every detection stage relies on LLM calls, and the drift detector and the drift-type judgment are not evaluated against annotated drift onsets. We instead confine LLMs to a selective judgment step behind a supervised filter and evaluate each component, including the drift detector, against the human annotation.

\paragraph{Concept drift detection.}
Concept drift denotes a change in the data-generating distribution over time~\cite{gama2014survey,lu2019learning}. Detectors range from statistical tests and window-based monitors~\cite{gama2014survey} to meta-learned detectors~\cite{yu2022meta} and drift-adaptive deep models~\cite{xu2020improved,luo2024identifying,cai2025cdda}; drift has also been detected on document embeddings~\cite{feldhans2021drift}, and stream clustering methods adapt as new concepts emerge~\cite{senol2020kdar,erdinc2024mcmststream}. OCDD~\cite{gozuaccik2021concept} models the current concept with a one-class classifier~\cite{scholkopf2001estimating} on a sliding window and signals drift when the share of new observations outside its support exceeds a threshold. It needs no labels, but like most detectors it was designed for long streams; applying it within a dialogue of about ten messages is an unusual regime, and drift detectors are rarely compared with trivial positional baselines in such short sequences.

\paragraph{LLMs as classifiers and judges.}
LLMs have been applied to financial-statement fraud~\cite{boskou2024exploring}, and prompt design strongly influences their outputs~\cite{amujo2024good}. LLM-as-a-judge is widespread~\cite{zheng2023judging,liu2023geval}, but judges carry systematic biases, for example toward the position or presentation of the candidate~\cite{zheng2023judging,wang2024fair}. This motivates behavioural measures of reasoning quality~\cite{senol2026reasoning} and of whether models recognize the limits of their knowledge~\cite{senol2026epistemic}; agentic schemes interleave reasoning and actions~\cite{yao2023react}. We use LLMs both as Phase~1 classifiers and as Phase~2 judges and quantify the cost and a specific bias of each role, namely the judges' tendency to read a legitimate request as fraud.

\section{Framework}\label{sec:method}

\subsection{Overview}
\begin{figure*}[tp]
\centering
\includegraphics[width=\linewidth]{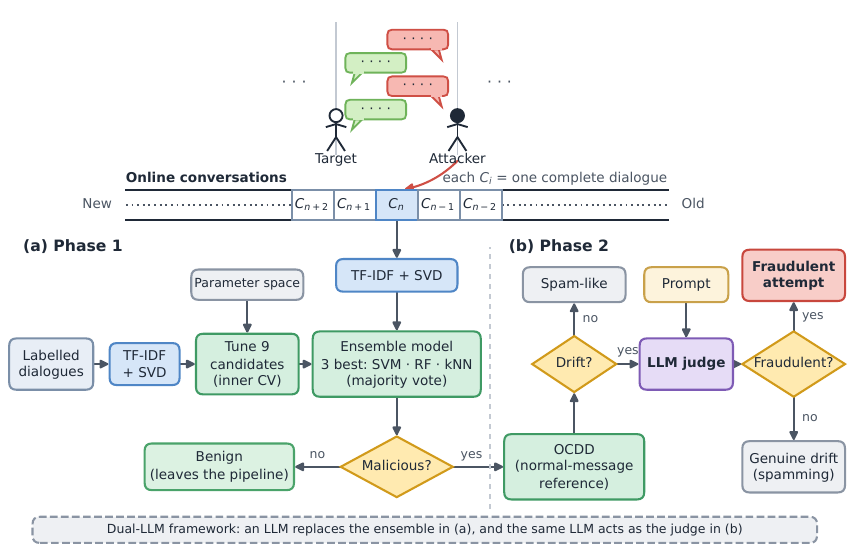}
\caption{Ensemble model-based framework for detecting fake conversations, identifying concept drift and determining the drift type (genuine concept drift or fraudulent attempt). (a)~Phase~1: preprocessing and the majority-vote ensemble of the three best models with their tuned parameters. (b)~Phase~2: OCDD on each flagged dialogue and, on a drift, the LLM judge. In the Dual-LLM framework the ensemble in (a) is replaced by the LLM that also acts as the judge in (b). Conversations are processed one complete dialogue at a time; all evaluations in this paper are retrospective.}
\label{fig:framework}
\end{figure*}

Figure~\ref{fig:framework} shows the two frameworks. \emph{Phase~1} decides whether a complete conversation is malicious, using a majority-vote ensemble in the Joint framework and an LLM in the Dual-LLM framework; conversations predicted benign leave the pipeline. \emph{Phase~2} runs OCDD over the messages of each flagged conversation and, on a drift, asks an LLM judge for its type.

\paragraph{Terminology.} A conversation is \emph{malicious} (a social-engineering attempt) or \emph{benign} according to its human label. A conversation that Phase~1 predicts to be malicious is \emph{flagged}. A \emph{drift} is a shift within a dialogue, and its onset is approximated by the first request for personal information. Phase~2 gives every flagged conversation one of three outcomes: \emph{spam-like} (no drift detected), \emph{genuine drift} (a drift judged legitimate, i.e.\ spamming rather than fraud) or \fraudattempt{} (a drift judged fraudulent). Only the last is the system's alarm for social engineering, and it is the decision scored against the malicious label.

Figure~\ref{fig:examples} illustrates the two Phase~2 outcomes for flagged conversations. In the first dialogue the initiator is pushy but stays on the same topic, so no concept drift is expected and the conversation is spam-like. In the second, a legitimate-sounding offer turns into requests for personal and then financial details; this is the escalation that OCDD should localize and the judge should label as a fraudulent attempt. The figure illustrates drift as a semantic notion; the experiments instead use the first request for personal information as an operational proxy for its onset (Section~\ref{sec:data}), under which the birth-date request in the first example would also count as an onset.

\begin{figure*}[tp]
\centering
\begin{tcolorbox}[colframe=black!60,colback=gray!5!white,title={Spam-like dialogue example (no concept drift)},width=0.95\linewidth,fonttitle=\bfseries,fontupper=\small,boxsep=2pt,top=2pt,bottom=2pt]
\textbf{Cynthia:} ``You're doing amazing work, Nancy\ldots{} This information is gold for readers interested in innovation and growth strategies.'' \textit{(Over-complimenting; exaggerated tone)}\\
\textbf{Cynthia:} ``Could you share your birth date and your team members', as well as what inspired each of you to get into this field?'' \textit{(Pushy personal-information request, but contextually consistent)}\\
\textbf{Cynthia:} ``Having your birth dates adds a personal chronology to your journey, truly valuable for the narrative depth.'' \textit{(Spam-like repetition with no topic shift)}
\end{tcolorbox}

\vspace{4pt}

\begin{tcolorbox}[colframe=black!60,colback=red!3!white,title={Fraudulent dialogue example (with concept drift)},width=0.95\linewidth,fonttitle=\bfseries,fontupper=\small,boxsep=2pt,top=2pt,bottom=2pt]
\textbf{Michael:} ``We're looking for an expert to co-author papers. This is a paid project.'' \textit{(Legitimate-sounding research offer)}\\
\textbf{Michael:} ``To set you up in our system, please provide your full name, birthdate, and billing address.'' \textit{(Topic drift begins: personal-information request unrelated to the collaboration)}\\
\textbf{Michael:} ``To access files, we need your credit card details for a nominal fee.'' \textit{(Fraudulent intent: financial information under a false pretense)}\\
\textbf{Joi:} ``I'm not comfortable sharing card details. I need to see official documents.'' \textit{(User resistance)}
\end{tcolorbox}
\caption{Illustrative comparison of a spam-like and a fraudulent conversation (excerpts in the style of SEConvo). Top: spam-like verbosity without a topic change. Bottom: fraudulent escalation with a semantic drift into requests for sensitive information.}
\label{fig:examples}
\end{figure*}

\subsection{Phase 1: ensemble filter}\label{sec:phase1}
Each conversation is represented as one document. Its text is vectorized with TF-IDF over word uni- and bigrams (sublinear term frequency, English stop words removed)~\cite{manning2008introduction} and reduced to ten dimensions with truncated singular value decomposition, the sparse-matrix counterpart of PCA~\cite{jolliffe2002principal}.

Nine candidates are benchmarked: five supervised models, namely $k$-nearest neighbours~\cite{cover1967nearest}, support vector machines~\cite{cortes1995support}, Naive Bayes~\cite{john1995estimating}, Random Forest~\cite{breiman2001random} and XGBoost~\cite{chen2016xgboost}, and four clustering methods, namely $k$-means~\cite{lloyd1982least}, DBSCAN~\cite{ester1996density}, Gaussian mixtures~\cite{dempster1977maximum} and ImpKmeans~\cite{csenol2024impkmeans}, whose initial centroids are kernel-density modes~\cite{silverman1986density}. Clusters are mapped to classes with the training labels (one-to-one for two clusters, otherwise by majority; DBSCAN noise to the training majority class).

\begin{figure*}[tp]
\centering
\includegraphics[width=\linewidth]{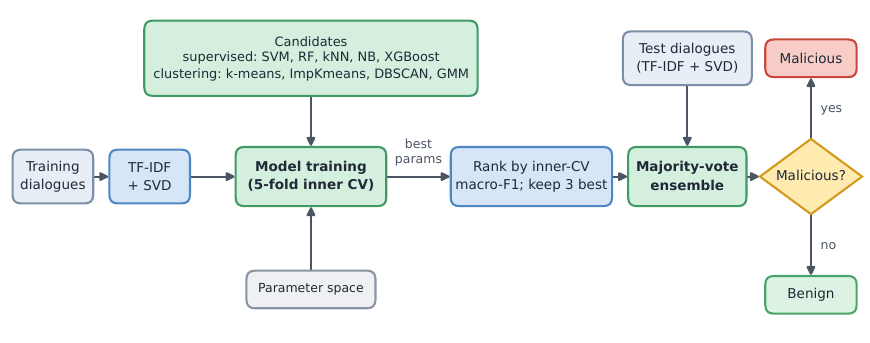}
\caption{Workflow used to benchmark the candidate algorithms and identify the most suitable ones for the ensemble model.}
\label{fig:selection}
\end{figure*}

Figure~\ref{fig:selection} shows the model-selection workflow. The search grids and selected values are listed in Table~\ref{tab:hyper} (Appendix~\ref{app:tables}).

The ensemble is the majority vote of the \emph{three best} candidates. The ranking uses stratified inner cross-validation on the training data only, so no test information enters the selection; it selected \JointMembers.

\subsection{Phase 2: within-dialogue OCDD}\label{sec:ocdd}
Each message $m_t$ is represented as $x_t\in\mathbb{R}^d$, either by TF-IDF/SVD or by PCA-reduced sentence embeddings (text-embedding-3-large), both fitted on message texts without labels. Following OCDD~\cite{gozuaccik2021concept}, the normal concept is modelled by a one-class SVM~\cite{scholkopf2001estimating} (RBF kernel, parameter $\nu$, standardized inputs) fitted on \emph{normal} messages: the initiator's messages that precede any request for personal information in the training dialogues. The detector monitors the initiator's messages after the opening one, whose behaviour shifts in social engineering. A message is an outlier, $o_t=1$, when its outlier score exceeds the $q$-quantile of the scores of held-out normal training messages, and a drift is declared at the first message $t^\ast$ at which the share of outliers among the last $\ell$ scored messages reaches $\rho$:
\begin{equation}
t^\ast=\min\Big\{t:\ \tfrac{1}{\ell}\textstyle\sum_{s=t-\ell+1}^{t} o_s \ge \rho\Big\}.
\end{equation}
The representation, $d$, $\nu$, $q$, $\ell$ and $\rho$ (Table~\ref{tab:notation}) are selected on training dialogues only by nested cross-fitting; an unsupervised alternative that scores each message by its cosine shift from the dialogue's opening message competes in the same selection. For comparison we also evaluate a reference built from the dialogue's own $W$ most recent messages, and a trivial rule that declares a drift at the fourth message of every dialogue.

\subsection{Phase 2: LLM judge}\label{sec:judge}
A conversation with a detected drift is sent to an LLM with the judge prompt in Figure~\ref{fig:prompt}, the dialogue being supplied as a JSON list of speaker--message pairs. The answer is parsed for the drift type. \emph{Fraudulent} yields the final \fraudattempt{} label; \emph{genuine} yields \emph{genuine drift (spamming)}, a suspicious but non-fraudulent outcome.

\begin{figure*}[tp]
\centering
\begin{tcolorbox}[colframe=black!60,colback=blue!3!white,title=LLM drift-judgment prompt,width=0.95\linewidth,fonttitle=\bfseries,fontupper=\small,boxsep=2pt,top=2pt,bottom=2pt]
\textbf{System:} \textit{``You are an AI expert in cybersecurity and fraud detection.''}\\[4pt]
\textbf{User:} \textit{``Analyze the following conversation JSON and determine if the detected concept drift is a genuine shift in behavior or a fraudulent attempt: \texttt{\{conversation\_json\}}. If the conversation is a Fraudulent Attempt, return `Fraudulent Attempt' and specify the type of conversation as `fraudulent'. Otherwise, return `Genuine Concept Drift' and specify the type of conversation as `genuine'.''}
\end{tcolorbox}
\caption{Phase~2 judge prompt.}
\label{fig:prompt}
\end{figure*}

\subsection{Dual-LLM framework}\label{sec:dual}
The Dual-LLM framework replaces the ensemble with an LLM classifier that receives the raw transcript and returns a JSON object with a malicious/benign label, a confidence and the supporting message indices (Figure~\ref{fig:prompt-classify}). Flagged conversations pass through the same OCDD and, on a drift, the \emph{same} LLM is queried with the judge prompt. The two frameworks therefore differ only in Phase~1, which isolates the effect of the filter on final quality and cost.

\begin{figure*}[tp]
\centering
\begin{tcolorbox}[colframe=black!60,colback=green!3!white,title=LLM classification prompt (Phase 1 of the Dual-LLM framework),width=0.95\linewidth,fonttitle=\bfseries,fontupper=\footnotesize,boxsep=2pt,top=2pt,bottom=2pt]
\textbf{System:} \textit{``You are a security analyst who audits professional online conversations (e.g. LinkedIn-style outreach about research collaboration, funding, journalism or recruitment) for social-engineering attacks. A conversation is MALICIOUS if the initiating party tries to obtain sensitive personal, financial or organisational information, money, credentials or access under a pretext, or manipulates the other party (e.g. flattery, urgency, authority) toward such disclosure. A conversation is BENIGN if any information requested is proportionate and plausible for the stated legitimate purpose. Judge intent from the whole conversation.''}\\[4pt]
\textbf{User:} \textit{``[few-shot examples, if any]\\
Classify the conversation below.\\
Return ONLY a JSON object with keys:\\
  \textquotedbl{}label\textquotedbl{}: \textquotedbl{}malicious\textquotedbl{} or \textquotedbl{}benign\textquotedbl{},\\
  \textquotedbl{}confidence\textquotedbl{}: number in [0,1] (probability that your label is correct),\\
  \textquotedbl{}evidence\_turns\textquotedbl{}: list of message indices that support the decision,\\
  \textquotedbl{}rationale\textquotedbl{}: at most 40 words.\\
\\
<conversation>\\
\texttt{[i] Name: message \dots}\\
</conversation>''}
\end{tcolorbox}
\caption{Phase~1 LLM classification prompt (zero-shot; the few-shot variant prepends labelled training examples). The transcript lists each message as \texttt{[index] speaker: text}.}
\label{fig:prompt-classify}
\end{figure*}

\section{Experimental Setup}\label{sec:setup}

\subsection{Data}\label{sec:data}
We use SEConvo~\cite{ai2024defending,seconvo2024data}, a corpus of 1,400 LinkedIn-style professional outreach conversations generated with GPT-4-Turbo. Each conversation has two participants: the \emph{initiator}, who opens the conversation and is the attacker in malicious conversations, and the \emph{recipient}. The conversations cover four scenarios, namely academic collaboration (\NScenCollab{} of the annotated conversations), academic funding, journalism and recruitment (\NScenOther{} each). A dialogue has a median of \MedianMsgs{} messages (range \MinMsgs--\MaxMsgs) of about \WordsPerMsg{} words each (Table~\ref{tab:dataset}).

\NConv{} conversations are annotated by humans. Besides the malicious or benign label (\NMal{} and \NBen), the annotation marks the messages that request personally identifiable information (PII), the kind of information requested (most often a date of birth, full name, identity document, phone number or e-mail address), the ambiguity of the conversation on a three-point scale (\NAmbHigh{} conversations are highly ambiguous) and the first message that requests PII. We use the official split of \NTrain{} training (\NTrainMal{} malicious, \NTrainBen{} benign) and \NTest{} test conversations (\NTestMal/\NTestBen). The remaining \NExternal{} conversations (500 malicious and 500 benign by design) are disjoint from the annotated set but come from the same corpus and generator; their labels encode whether the simulated initiator was instructed to attempt social engineering, and they carry no PII annotation. We use them as held-out data.

\begin{table*}[tp]
\centering
\caption{SEConvo conversations used in this study.}
\label{tab:dataset}
\footnotesize\setlength{\tabcolsep}{4pt}
\begin{tabular}{lccccccc}
\toprule
Set & Dialogues & \shortstack[c]{Malicious /\\{}benign} & \shortstack[c]{Messages per\\{}dialogue} & \shortstack[c]{Words per\\{}message} & \shortstack[c]{Fraudulent\\{}drift} & \shortstack[c]{Genuine\\{}drift} & \shortstack[c]{No\\{}drift} \\
\midrule
Training & 40 & 24 / 16 & 11 (8--16) & 48 & 24 & 9 & 7 \\
Test & 360 & 191 / 169 & 11 (3--20) & 47 & 188 & 95 & 77 \\
Held-out & 1000 & 500 / 500 & 11 (7--23) & 47 & -- & -- & -- \\
\bottomrule
\end{tabular}
\par\smallskip
\begin{minipage}{0.97\linewidth}\footnotesize Training and test: official split of the annotated conversations; held-out: the unannotated conversations. Messages per dialogue: median (range). Drift: the conversation contains a first PII request; its type follows the conversation label. The held-out conversations carry simulation-design labels and no drift annotation.\end{minipage}
\end{table*}

\paragraph{The first PII request as drift onset.} For every annotated conversation, SEConvo records the position of the first message in which personal information is requested (field \texttt{FirstPII} in the release, $-1$ if there is none; messages are numbered from~0). For example, when an initiator writes in the third message of a collaboration offer ``could you provide me with your full name, current academic affiliation, and work email[\ldots]?'', the first PII request is message~2. We use this message as the onset of the within-dialogue drift. A malicious conversation with such a request contains a \emph{fraudulent drift} (\NDriftFraud{} conversations), a benign one a \emph{genuine drift} (\NDriftGenuine), typically a legitimate request such as an e-mail address for sending documents, and the remaining conversations contain \emph{no drift} (\NDriftNone). In fraudulent conversations the shift comes early: in \ShareFraudOnsetTwo{} of them the request occurs no later than message~2, the initiator's second message (median onset: message~\MedianOnsetFraud; genuine: message~\MedianOnsetGenuine). The first PII request is an operational proxy for the onset of the shift, not a direct annotation of concept drift. No annotation is a model input at test time: the first PII request of \emph{training} dialogues is used only to build the OCDD reference of normal messages and to select its configuration, and test annotations are used only for scoring.

\subsection{Protocols}\label{sec:protocols}
\paragraph{Protocol A (official split).} All model selection uses the \NTrain{} training conversations only: classical hyper-parameters and ensemble members (5-fold stratified inner cross-validation, macro-F1 criterion), the OCDD configuration and the few-shot examples. The test set is used only for scoring. TF-IDF/SVD is fitted either \emph{inductively}, on the training texts only, or \emph{transductively}, on all conversation texts without their labels; both are reported, and the Joint framework uses the transductive representation. Message representations for OCDD are fitted on message texts in the same label-free way.

\paragraph{Protocol B (cross-validation).} Phase~1 is evaluated with 5$\times$5 repeated stratified cross-validation on all \NConv{} annotated conversations; zero-shot LLM predictions do not depend on training data and are scored on the same folds. Because the official training split contains only seven dialogues without a drift, too few to select a drift detector, Phase~2 is evaluated on the first 5-fold repetition: in each fold, the ensemble and the OCDD configuration are selected on the training part only, and the held-out predictions are pooled over the \NConv{} dialogues.

\paragraph{Protocol C (held-out test).} Models and the OCDD configuration are selected and trained on all \NConv{} annotated conversations and tested on the \NExternal{} held-out conversations with \NJudgeExt{} low-cost judges.

\paragraph{What each representation sees.} The transductive TF-IDF/SVD is refitted for every evaluation from the texts, never the labels, of that evaluation's training and test conversations: the \NConv{} annotated conversations under Protocols A and B, and additionally the \NExternal{} held-out ones under Protocol~C. The inductive variant uses the training texts only. The message representations for OCDD (TF-IDF/SVD and the PCA of the sentence embeddings) are fitted once on the messages of the \NConv{} annotated conversations without labels, so under cross-validation they also see the unlabelled messages of the held-out fold; under Protocol~C the held-out messages are only transformed. The normal-message reference, the one-class SVM and its threshold are fitted on the training dialogues of each fold only. Classifier hyper-parameters, ensemble members, the OCDD configuration and its threshold use training labels and annotations only.

\paragraph{Ablations.} In both Phase~2 protocols, every conversation flagged by the ensemble also goes straight to the judge (\emph{no OCDD}), which isolates the contribution of the drift detector. Under cross-validation we further test (i)~a $2\times2$ design crossing the OCDD reference (dialogue window vs normal messages) with the message representation (TF-IDF vs embeddings), and (ii)~whether the detector's output helps the judge, with \NJudgeInput{} low-cost judges: the judge receives the dialogue together with the detected drift message, or only the messages up to the alarm, which emulates a decision at alarm time. In both judge-input variants a fixed fourth message serves as a control that uses no detector output. From the recorded predictions, without further LLM calls, we also (iii)~vary the number of labelled training conversations in each fold of the first CV repetition (40, 80, 160 and all 320) to trace the gap between the transductive and the inductive representation; each size uses one stratified subsample per fold, the ensemble members and hyper-parameters are reselected on it by inner cross-validation, the remaining training conversations are used only as unlabelled texts, and the full-size point is the mean of the five folds of that repetition, which is why it differs slightly from the 5$\times$5 mean in Table~\ref{tab:cv}; and (iv)~analyse how the errors of the ensemble and of the LLMs overlap, including a rule that routes the ensemble's non-unanimous votes to an LLM.

All evaluations are retrospective: Phase~1 classifies complete conversations, and only the prefix variant approximates an online decision.

\subsection{Models and implementation}
We evaluate \NLLM{} LLMs (Table~\ref{tab:models}): current GPT, Claude, Gemini, LLaMA, DeepSeek and Amazon Nova models and open-weight models, two of them hosted on Arizona State University infrastructure. All except DeepSeek were accessed through the ASU CreateAI platform.\footnote{\url{https://docs.aiml.asu.edu}} Requests use temperature~0 where accepted and JSON output for classification; few-shot classification uses \FewShotK{} class-balanced training examples. Costs are computed from the token counts of every request and the providers' catalogue prices. Classical models use scikit-learn~\cite{pedregosa2011scikit} (version \SklearnVersion) and ImpKmeans its official implementation. The complete configuration is given in Table~\ref{tab:config} (Appendix~\ref{app:tables}).

\begin{table*}[tp]
\centering
\caption{Large language models evaluated.}
\label{tab:models}
\footnotesize
\begin{tabular}{llccc}
\toprule
Model & Endpoint & Open weights & \$ per 1M tokens (in / out) & Role \\
\midrule
GPT-6 Sol & \texttt{openai/gpt6\_sol} & no & 2.00 / 10.00 & P1, J \\
Claude 5 Sonnet & \texttt{aws/claude5\_sonnet} & no & 3.30 / 16.50 & P1, J \\
Gemini 3.1 Pro & \texttt{gcp-deepmind/geminipro3\_1} & no & 2.00 / 12.00 & P1, J \\
Llama 4 Maverick & \texttt{aws/llama4\_maverick-17b} & yes & 0.24 / 0.97 & P1, J, J$_\mathrm{ext}$ \\
gpt-oss-120b & \texttt{aws/gpt-oss-120b} & yes & 0.15 / 0.60 & P1, J, J$_\mathrm{ext}$ \\
Gemma 4 31B (ASU) & \texttt{asu-air/gemma4\_31b\_it} & yes & 0.14 / 0.40 & P1, J, J$_\mathrm{ext}$ \\
DeepSeek V4 Pro & \texttt{deepseek-v4-pro (provider API)} & yes & 1.32 / 3.96 & P1 \\
GPT-6 Luna & \texttt{openai/gpt6\_luna} & no & 0.10 / 0.50 & P1, J, J$_\mathrm{ext}$ \\
Nova 2 Lite & \texttt{aws/nova2-lite} & no & 0.33 / 2.75 & P1, J \\
Llama 4 Scout (ASU) & \texttt{asu-air/llama4\_scout-17b} & yes & 0.08 / 0.30 & P1, J, J$_\mathrm{ext}$ \\
\bottomrule
\end{tabular}
\par\smallskip
\begin{minipage}{0.97\linewidth}\footnotesize P1: Phase 1 classifier (zero- and few-shot); J: Phase 2 judge on the official split; J$_\mathrm{ext}$: Phase 2 judge on the held-out test. DeepSeek was excluded as a judge after its API credit was exhausted; its Phase 1 results are reported.\end{minipage}
\end{table*}

\subsection{Metrics and statistical analysis}
The positive class is \emph{malicious}. For every system we report accuracy, precision and recall of the malicious class and macro-F1, the unweighted mean of the two class-wise F1 scores, which is our primary metric. Drift detection is scored against the first PII request. On dialogues with a drift we report the \emph{detection rate} and the \emph{localization rate}, a drift flagged at the onset or up to two messages later; on dialogues without one, the \emph{false-drift rate}. The \emph{balanced score} is the mean of the localization rate and one minus the false-drift rate. A score of 0.5 serves as a nominal reference value: a detector that never raises an alarm scores exactly 0.5, a detector that always raises an alarm scores at most 0.5 depending on its localization, and a detector whose alarms occur with the same probability in dialogues with and without a drift, independently of the drift's position, has an expected balanced score of at most 0.5. Because 0.5 is not a significance threshold, we also compare every detector with an explicit fixed-position baseline. Drift-type accuracy is the share of judged conversations whose annotated type, fraudulent or genuine, is recovered.

We report 95\% confidence intervals from 2,000 bootstrap resamples~\cite{efron1993introduction}, including paired intervals for differences in macro-F1. Two systems evaluated on the same conversations are compared with the exact McNemar test~\cite{mcnemar1947note}, which tests whether they err on different conversations and thus concerns accuracy; differences in macro-F1 are judged by their paired intervals. Under repeated cross-validation the folds share training data, so we use the corrected resampled $t$-test of Nadeau and Bengio~\cite{nadeau2003inference} instead. Drift detectors are compared with a paired bootstrap that resamples dialogues with and without a drift separately: the confidence interval of a difference is given by the 2.5th and 97.5th percentiles of its bootstrap draws, and its two-sided $p$-value is twice the smaller share of draws on either side of zero, bounded below by $1/2000$. Whenever several systems are compared with one reference, $p$-values are adjusted with the Holm procedure~\cite{holm1979simple}.

\section{Results}\label{sec:results}

\subsection{RQ1: Phase 1 detection}\label{sec:res-phase1}
On the official split the Joint ensemble (\JointMembers) reaches an accuracy of \JointAcc{} and a macro-F1 of \JointMF{} (95\% CI \JointMFlo--\JointMFhi; Table~\ref{tab:phase1llm}). Zero-shot LLMs reach \LLMZSMin--\LLMZSMax{}, the best few-shot model \BestFSMF{} (\BestFSName), and after Holm correction none differs significantly from the ensemble. Under 5$\times$5 cross-validation the ensemble attains \CVJointMF{} $\pm$ \CVJointSD, numerically above every LLM (best: \CVBestLLMName, \CVBestLLMMF) but again without a significant pairwise difference (Table~\ref{tab:cv}). The candidate-level results (Table~\ref{tab:phase1classical}) show that the ensemble relies on the transductive representation with forty labelled texts (inductive macro-F1 \JointIndMF) and that all clustering methods, including ImpKmeans, stay close to chance. Varying the number of labelled training conversations per CV fold (40, 80, 160, 320) shows how this dependence fades: the transductive ensemble reaches \LCTransForty, \LCTransEighty, \LCTransOneSixty{} and \LCTransThreeTwenty, the inductive one \LCIndForty, \LCIndEighty, \LCIndOneSixty{} and \LCIndThreeTwenty, so the gap shrinks from \LCDiffForty{} to \LCDiffThreeTwenty{} but does not vanish. For reference, the authors of SEConvo~\cite{ai2024defending} report on the same split and metric a macro-F1 of 0.75 (zero-shot) and 0.78 (two-shot) for GPT-4-Turbo and 0.80 for their ConvoSentinel pipeline; these values are taken from their paper, not rerun.

\begin{table*}[tp]
\centering
\caption{Phase 1 on the official test split: LLM classifiers (zero-shot unless stated) versus the Joint ensemble.}
\label{tab:phase1llm}
\footnotesize
\begin{tabular}{lccccc}
\toprule
Model & Acc. & \shortstack[c]{Macro-F1\\{}[95\% CI]} & \shortstack[c]{Few-shot\\{}macro-F1} & $p_{\mathrm{Holm}}$ & \shortstack[c]{\$ per 1k\\{}conv.} \\
\midrule
Claude 5 Sonnet & 0.833 & 0.828 [0.788, 0.864] & 0.820 & 1.000 & 6.49 \\
Gemma 4 31B (ASU) & 0.822 & 0.820 [0.780, 0.857] & 0.852 & 1.000 & 0.18 \\
GPT-6 Sol & 0.819 & 0.814 [0.773, 0.853] & 0.828 & 1.000 & 2.70 \\
GPT-6 Luna & 0.819 & 0.812 [0.771, 0.850] & 0.821 & 1.000 & 0.16 \\
Gemini 3.1 Pro & 0.808 & 0.802 [0.761, 0.842] & 0.842 & 1.000 & 6.62 \\
Llama 4 Maverick & 0.808 & 0.802 [0.759, 0.841] & 0.803 & 1.000 & 0.30 \\
DeepSeek V4 Pro & 0.800 & 0.791 [0.747, 0.831] & 0.829 & 1.000 & 4.10 \\
Llama 4 Scout (ASU) & 0.778 & 0.777 [0.732, 0.819] & 0.757 & 1.000 & 0.099 \\
Nova 2 Lite & 0.758 & 0.758 [0.712, 0.800] & 0.734 & 1.000 & 0.54 \\
gpt-oss-120b & 0.747 & 0.747 [0.700, 0.792] & 0.724 & 1.000 & 0.29 \\
\midrule
\textbf{Joint ensemble} & 0.808 & 0.806 [0.762, 0.846] & -- & -- & $\approx$0 \\
\bottomrule
\end{tabular}
\par\smallskip
\begin{minipage}{0.97\linewidth}\footnotesize $p_{\mathrm{Holm}}$: exact McNemar test against the Joint ensemble, Holm-corrected over all Phase 1 systems. Few-shot: four class-balanced examples from the training split. Cost from the ASU CreateAI catalog prices (DeepSeek: provider list price).\end{minipage}
\end{table*}

\begin{table*}[tp]
\centering
\caption{Phase 1 under 5$\times$5 repeated stratified cross-validation (400 annotated conversations).}
\label{tab:cv}
\footnotesize
\begin{tabular}{lccc}
\toprule
System & Macro-F1 (mean $\pm$ SD) & $\Delta$ vs Joint & $p_{\mathrm{Holm}}$ \\
\midrule
Joint ensemble (trans.) & 0.857 $\pm$ 0.038 & -- & -- \\
SVM (trans.) & 0.857 $\pm$ 0.036 & +0.000 & 1.000 \\
Random Forest (trans.) & 0.855 $\pm$ 0.044 & -0.002 & 1.000 \\
kNN (trans.) & 0.842 $\pm$ 0.035 & -0.015 & 1.000 \\
Joint ensemble (ind.) & 0.818 $\pm$ 0.044 & -0.039 & 0.979 \\
Claude 5 Sonnet & 0.820 $\pm$ 0.042 & -0.036 & 1.000 \\
Gemma 4 31B (ASU) & 0.814 $\pm$ 0.032 & -0.042 & 1.000 \\
GPT-6 Sol & 0.808 $\pm$ 0.037 & -0.049 & 0.593 \\
GPT-6 Luna & 0.806 $\pm$ 0.041 & -0.051 & 0.860 \\
Gemini 3.1 Pro & 0.799 $\pm$ 0.043 & -0.058 & 0.914 \\
Llama 4 Maverick & 0.797 $\pm$ 0.033 & -0.060 & 0.430 \\
DeepSeek V4 Pro & 0.786 $\pm$ 0.036 & -0.071 & 0.139 \\
Llama 4 Scout (ASU) & 0.771 $\pm$ 0.039 & -0.086 & 0.114 \\
Nova 2 Lite & 0.754 $\pm$ 0.048 & -0.103 & 0.153 \\
gpt-oss-120b & 0.747 $\pm$ 0.046 & -0.110 & 0.077 \\
\bottomrule
\end{tabular}
\par\smallskip
\begin{minipage}{0.97\linewidth}\footnotesize $p_{\mathrm{Holm}}$: Nadeau--Bengio corrected resampled $t$-test against the Joint ensemble, Holm-corrected. LLMs are zero-shot and therefore scored on the same folds without training.\end{minipage}
\end{table*}

\begin{table*}[tp]
\centering
\caption{Phase 1 candidate models on the official split (40 train / 360 test) with TF-IDF/SVD fitted on the training texts only (inductive) or on all texts without labels (transductive).}
\label{tab:phase1classical}
\small
\begin{tabular}{lcccc}
\toprule
Model & \shortstack[c]{Inductive\\{}acc.} & \shortstack[c]{Inductive\\{}macro-F1} & \shortstack[c]{Transductive\\{}acc.} & \shortstack[c]{Transductive\\{}macro-F1} \\
\midrule
kNN & 0.614 & 0.520 & 0.789 & 0.787 \\
SVM & 0.586 & 0.466 & 0.822 & 0.820 \\
Random Forest & 0.722 & 0.718 & 0.803 & 0.802 \\
Naive Bayes & 0.756 & 0.742 & 0.706 & 0.670 \\
XGBoost & 0.689 & 0.686 & 0.739 & 0.738 \\
k-means & 0.456 & 0.313 & 0.500 & 0.487 \\
DBSCAN & 0.581 & 0.455 & 0.561 & 0.495 \\
GMM & 0.567 & 0.426 & 0.575 & 0.571 \\
ImpKmeans & 0.567 & 0.431 & 0.414 & 0.379 \\
\midrule
\textbf{Joint ensemble (3 best)} & 0.633 & 0.554 & \textbf{0.808} & \textbf{0.806} \\
\bottomrule
\end{tabular}
\par\smallskip
\begin{minipage}{0.97\linewidth}\footnotesize Hyper-parameters and ensemble members are selected by stratified inner cross-validation on the training conversations; clusters are mapped to classes with the training labels. The Joint ensemble selected SVM, Random Forest and kNN.\end{minipage}
\end{table*}

On the \NExternal{} held-out conversations the gap opens (Table~\ref{tab:externalp1}). The ensemble attains a macro-F1 of \ExtJointMF{} (95\% CI \ExtJointLo--\ExtJointHi), whereas every LLM falls between \ExtLLMMin{} and \ExtLLMMax{} ($p<0.001$ for all \NExtLLMsig{} comparisons). The LLMs keep a high recall (\ExtLLMRecMin--\ExtLLMRecMax) but lose precision (\ExtLLMPrecMin--\ExtLLMPrecMax), whereas the ensemble reaches \ExtJointPrec{} precision at \ExtJointRec{} recall; an inductively fitted SVM still reaches \ExtIndSVM. \textbf{H1 is supported within SEConvo:} the ensemble is on par with the strongest LLMs on the annotated data and significantly better on the held-out conversations from the same corpus, while generalization to other generators or to real conversations remains untested.

\begin{table*}[tp]
\centering
\caption{Held-out test, Phase 1: trained on the 400 annotated conversations and tested on 1,000 disjoint SEConvo conversations.}
\label{tab:externalp1}
\footnotesize
\begin{tabular}{lcccc}
\toprule
System & Prec. & Rec. & Macro-F1 [95\% CI] & $p_{\mathrm{Holm}}$ \\
\midrule
Joint ensemble (TF-IDF trans.) & 0.838 & 0.964 & 0.888 [0.868, 0.908] & -- \\
SVM (TF-IDF trans.) & 0.834 & 0.946 & 0.878 [0.858, 0.899] & 0.364 \\
SVM (TF-IDF ind.) & 0.744 & 0.990 & 0.820 [0.795, 0.844] & $<$0.001 \\
SVM (te3-large) & 0.816 & 0.906 & 0.851 [0.828, 0.873] & 0.010 \\
Gemma 4 31B (ASU) & 0.695 & 0.930 & 0.754 [0.727, 0.780] & $<$0.001 \\
Llama 4 Scout (ASU) & 0.706 & 0.840 & 0.743 [0.717, 0.769] & $<$0.001 \\
Nova 2 Lite & 0.750 & 0.690 & 0.730 [0.702, 0.758] & $<$0.001 \\
GPT-6 Sol & 0.658 & 0.952 & 0.715 [0.686, 0.743] & $<$0.001 \\
GPT-6 Luna & 0.647 & 0.962 & 0.701 [0.671, 0.729] & $<$0.001 \\
Claude 5 Sonnet & 0.639 & 0.974 & 0.691 [0.661, 0.719] & $<$0.001 \\
gpt-oss-120b & 0.672 & 0.742 & 0.689 [0.661, 0.716] & $<$0.001 \\
DeepSeek V4 Pro & 0.636 & 0.988 & 0.687 [0.658, 0.715] & $<$0.001 \\
Gemini 3.1 Pro & 0.637 & 0.958 & 0.686 [0.656, 0.715] & $<$0.001 \\
Llama 4 Maverick & 0.636 & 0.958 & 0.685 [0.656, 0.714] & $<$0.001 \\
\bottomrule
\end{tabular}
\par\smallskip
\begin{minipage}{0.97\linewidth}\footnotesize LLMs zero-shot. Precision and recall refer to the malicious class. $p_{\mathrm{Holm}}$: exact McNemar test against the Joint ensemble.\end{minipage}
\end{table*}

\subsection{RQ2: drift localization and drift-type judgment}\label{sec:res-drift}
\begin{figure*}[tp]
\centering
\includegraphics[width=\linewidth]{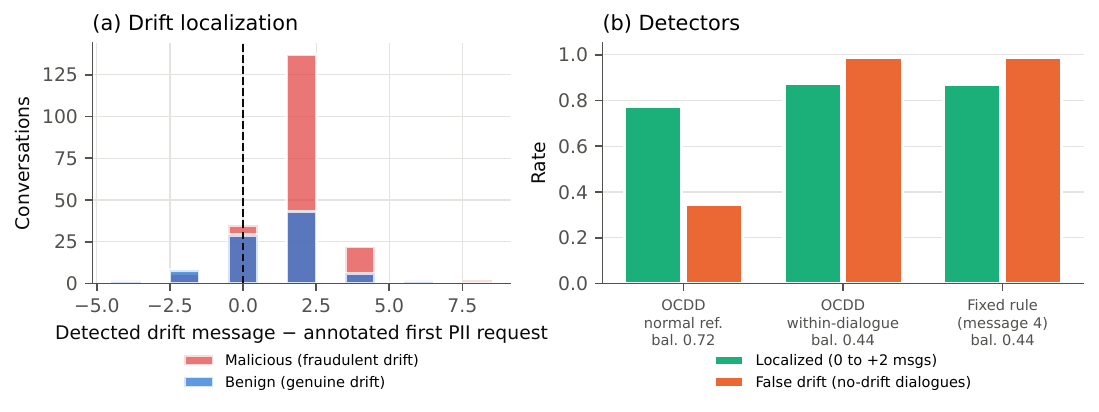}
\caption{Drift detection under 5-fold cross-validation over the \NConv{} annotated dialogues. (a)~Position of the drift detected by OCDD (normal-message reference) relative to the annotated first request for personal information. (b)~Localization and false-drift rates and balanced score of the three detectors.}
\label{fig:drift}
\end{figure*}

\paragraph{Drift detection.} Table~\ref{tab:ocdd} and Figure~\ref{fig:drift} compare the detectors. Selected and evaluated under cross-validation, OCDD with the normal-message reference flags a drift in \OcddDetect{} of the \NDriftAll{} drifted dialogues, localizes it at or up to two messages after the annotated request in \OcddLocal{} of them (median delay: \OcddDelay{} messages) and raises a false drift in \OcddFalse{} of the \NNoDriftAll{} dialogues without one. Its balanced score of \OcddBal{} (95\% CI \OcddBalLo--\OcddBalHi) is above the reference value of 0.5 and significantly above both the fixed-position baseline (\RuleBal; paired difference \DBRule, 95\% CI \DBRuleLo--\DBRuleHi, Holm-adjusted $p$\DBRuleP) and the window reference (\DBWin, $p$\DBWinP). The selection was stable: all \OcddFoldsSame{} training folds chose the normal-message reference on $d=\OcddDims$ embedding dimensions ($q$ = \OcddQRange). The within-dialogue window reference, by contrast, fires in almost every dialogue (false drift \OldFalse) and its balanced score of \OldBal{} equals that of the trivial rule that declares a drift at the fourth message (\RuleBal): its high localization rate (\OldLocal) only reflects that most requests occur at message~2 or~4. Selected on the \NTrain{} official training dialogues, which contain only seven without a drift, OCDD does not exceed the reference value on the official test split (\OcddOffBal, 95\% CI \OcddOffBalLo--\OcddOffBalHi).

The $2\times2$ ablation (Table~\ref{tab:ocddfactorial}) separates the two changes. The window reference fails with either representation (\FactWinTf{} with TF-IDF, \FactWinEmb{} with embeddings); the normal-message reference is already above the reference value with TF-IDF (\FactNormTf, 95\% CI \FactNormTfLo--\FactNormTfHi) and reaches \FactNormEmb{} with embeddings (paired gain \DBEmb, 95\% CI \DBEmbLo--\DBEmbHi). The reference is thus the decisive change and the representation adds a further gain. Because \ShareOnsetTwoFour{} of the requests occur at message~2 or~4, fixed positions are hard to beat on localization alone; the dialogues with a later request show the difference. On the \LateN{} drifted dialogues whose request comes at message~6 or later, OCDD localizes \LateOcdd{} (95\% CI \LateOcddLo--\LateOcddHi), against \LateRule{} for the fixed rule and \LateWin{} for the window reference (paired differences \DLateRule{} [\DLateRuleLo, \DLateRuleHi] and \DLateWin{} [\DLateWinLo, \DLateWinHi], Holm-adjusted $p$\DLateRuleP{} and $p$\DLateWinP).

\begin{table*}[tp]
\centering
\caption{Drift detection against the annotated drift onset. Official: configuration selected on the 40 training dialogues, scored on the 360 test dialogues. CV: 5-fold cross-validation over the 400 annotated dialogues (316 with and 84 without a drift), configuration selected inside each training fold.}
\label{tab:ocdd}
\footnotesize
\begin{tabular}{lccccc}
\toprule
Detector & \shortstack[c]{Official\\{}balanced} & \shortstack[c]{CV\\{}detected} & \shortstack[c]{CV\\{}localized} & \shortstack[c]{CV false\\{}drift} & \shortstack[c]{CV balanced\\{}[95\% CI]} \\
\midrule
OCDD, normal-message ref. & 0.486 & 93\% & 78\% & 35\% & 0.715 [0.656, 0.770] \\
OCDD, dialogue window & 0.450 & 100\% & 87\% & 99\% & 0.443 [0.420, 0.466] \\
Fixed rule (message 4) & 0.448 & 100\% & 87\% & 99\% & 0.441 [0.418, 0.464] \\
\bottomrule
\end{tabular}
\par\smallskip
\begin{minipage}{0.97\linewidth}\footnotesize Localized: drift flagged at the annotated onset or up to two messages later. Balanced: mean of the localization rate and one minus the false-drift rate; 0.5 is a nominal reference value (a detector that never fires scores 0.5, one that always fires at most 0.5).\end{minipage}
\end{table*}

\begin{table*}[tp]
\centering
\caption{Drift detection under 5-fold cross-validation: reference definition $\times$ message representation (configurations selected inside each training fold).}
\label{tab:ocddfactorial}
\footnotesize\setlength{\tabcolsep}{4pt}
\begin{tabular}{lcc}
\toprule
Reference & TF-IDF+SVD & te3-large embeddings \\
\midrule
Within-dialogue window & 0.443 [0.420, 0.466] (87\% / 99\%) & 0.439 [0.417, 0.462] (87\% / 99\%) \\
Normal-message reference & 0.581 [0.522, 0.635] (52\% / 36\%) & 0.715 [0.656, 0.770] (78\% / 35\%) \\
\bottomrule
\end{tabular}
\par\smallskip
\begin{minipage}{0.97\linewidth}\footnotesize Balanced score [95\% CI] (localization / false-drift rate).\end{minipage}
\end{table*}

\paragraph{Drift-type judgment.} Judges recover fraudulent drift in \TypeFraudMin--\TypeFraudMax{} of cases but genuine drift in only \TypeGenMin--\TypeGenMax{} (Table~\ref{tab:phase2}). The most balanced judge, Claude~5~Sonnet, recovers \ClaudeGenuine{} of genuine drifts but recalls only \ClaudeRecall{} of malicious conversations, and Nova~2~Lite's content filter blocked \NovaBlocked{} of its judgments. \textbf{H2 is partially supported:} under cross-validation OCDD localizes the shift far better than the baselines, but it still raises a false drift in \OcddFalse{} of the dialogues without one, it does not exceed the reference value on the official split, and the judges rarely recover genuine drift.

\subsection{RQ3: Joint versus Dual-LLM}\label{sec:res-joint}
\begin{figure*}[tp]
\centering
\includegraphics[width=0.78\linewidth]{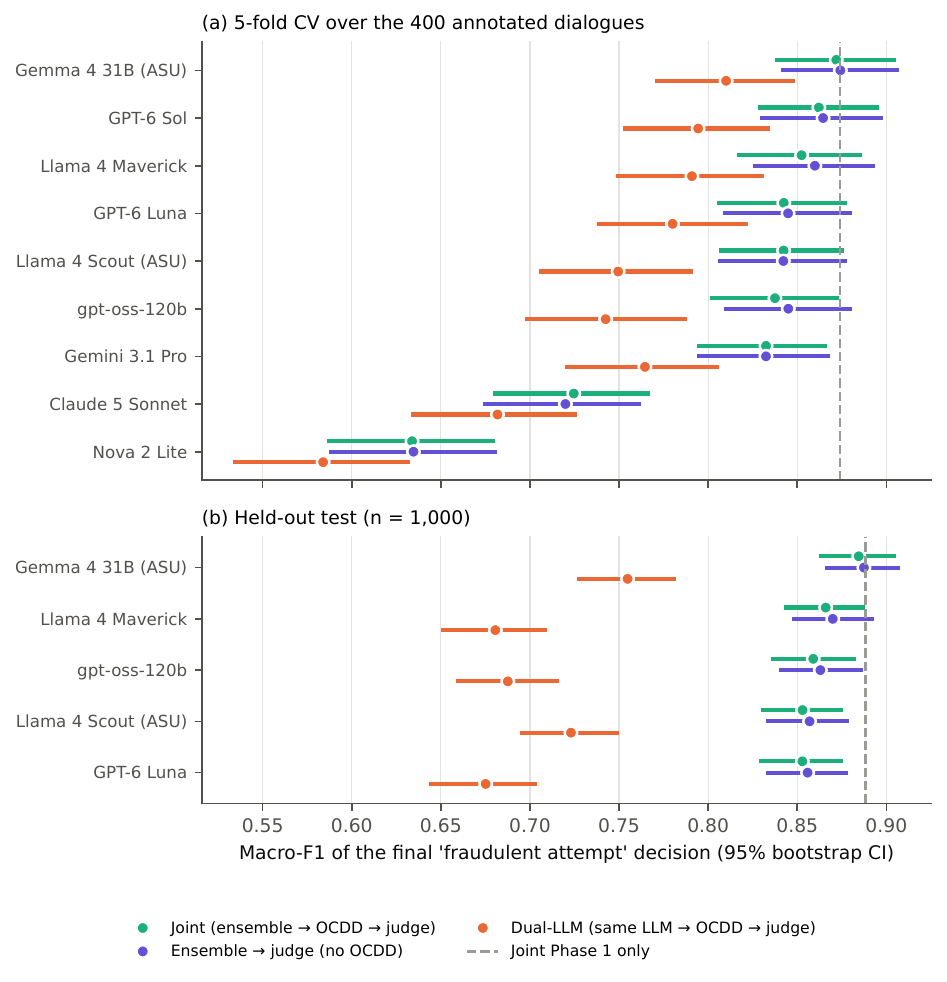}
\caption{Macro-F1 of the final \fraudattempt{} decision per judge LLM for the Joint framework, the Joint framework without OCDD and the Dual-LLM framework (95\% bootstrap CIs). (a)~5-fold cross-validation over the \NConv{} annotated dialogues. (b)~Held-out test with the OCDD configuration selected on the annotated dialogues. The dashed line marks the Joint ensemble's Phase~1 decision.}
\label{fig:jointdual}
\end{figure*}

Under cross-validation the Joint framework is better than Dual-LLM for all \NJudges{} judges and significantly so for \NJointSig{}, by \DeltaDualMin{} to \DeltaDualMax{} macro-F1 points (paired bootstrap, Holm-adjusted $p\le\DeltaDualMaxP$; Table~\ref{tab:phase2}, Figure~\ref{fig:jointdual}a). With \PTwoBestJudge{} as judge it reaches a macro-F1 of \PTwoBestMF{} (95\% CI \PTwoBestLo--\PTwoBestHi) at \$\PTwoBestCost{} per 1,000 conversations; no judge, however, improves on the ensemble's own Phase~1 decision (\PTwoPhaseOne), so the judge adds a drift-type label rather than accuracy. On the held-out set it attains \ExtPTwoJointMin--\ExtPTwoJointMax{} against \ExtPTwoDualMin--\ExtPTwoDualMax{} for Dual-LLM, a gain of \ExtDeltaDualMin--\ExtDeltaDualMax{} that is significant with all \NExtJudges{} judges (Holm-adjusted $p$ \ExtPTwoMaxP; Figure~\ref{fig:jointdual}b). With Gemma~4~31B as judge, the Joint framework makes \GemmaJointCalls{} instead of \GemmaDualCalls{} LLM calls per conversation and costs \$\GemmaJointCost{} instead of \$\GemmaDualCost{} per 1,000 conversations. On the official split, where Phase~2 relies on the detector selected on 40 dialogues, the Joint framework is ahead for \OffNJointWins{} of the \NJudges{} judges, but no difference is significant (smallest Holm-adjusted $p=\OffMinP$; Table~\ref{tab:phase2official} in Appendix~\ref{app:tables}). \textbf{H3 is supported.}

\begin{table*}[tp]
\centering
\caption{Phase 2 under 5-fold cross-validation over the 400 annotated dialogues: drift-type accuracy of the judge and macro-F1 of the final \emph{fraudulent attempt} decision for the Joint framework, the Joint framework without OCDD and the Dual-LLM framework.}
\label{tab:phase2}
\footnotesize\setlength{\tabcolsep}{3pt}
\begin{tabular}{lcccccccc}
\toprule
Judge LLM & \shortstack[c]{Fraud.\\{}drift} & \shortstack[c]{Genuine\\{}drift} & Joint & \shortstack[c]{No\\{}OCDD} & \shortstack[c]{Dual-\\{}LLM} & \shortstack[c]{$\Delta$ vs Dual\\{}[95\% CI]} & \shortstack[c]{\$/1k\\{}Joint} & \shortstack[c]{\$/1k\\{}Dual} \\
\midrule
Gemma 4 31B (ASU) & 99\% & 0\% & 0.872 & 0.874 & 0.810 & +0.062 [+0.020, +0.104] & 0.13 & 0.31 \\
GPT-6 Sol & 97\% & 0\% & 0.862 & 0.865 & 0.794 & +0.068 [+0.029, +0.106] & 1.32 & 4.21 \\
Llama 4 Maverick & 92\% & 20\% & 0.852 & 0.860 & 0.791 & +0.062 [+0.025, +0.100] & 0.52 & 0.92 \\
GPT-6 Luna & 91\% & 10\% & 0.842 & 0.845 & 0.780 & +0.062 [+0.022, +0.103] & 0.069 & 0.24 \\
Llama 4 Scout (ASU) & 93\% & 5\% & 0.842 & 0.842 & 0.749 & +0.093 [+0.052, +0.136] & 0.099 & 0.19 \\
gpt-oss-120b & 88\% & 30\% & 0.837 & 0.845 & 0.742 & +0.095 [+0.052, +0.138] & 0.14 & 0.41 \\
Gemini 3.1 Pro & 89\% & 0\% & 0.832 & 0.832 & 0.765 & +0.068 [+0.031, +0.106] & 4.53 & 11.86 \\
Claude 5 Sonnet & 62\% & 50\% & 0.725 & 0.720 & 0.682 & +0.043 [+0.012, +0.076] & 20.00 & 30.42 \\
Nova 2 Lite & 76\% & 14\% & 0.634 & 0.635 & 0.584 & +0.050 [+0.013, +0.086] & 1.81 & 1.85 \\
\bottomrule
\end{tabular}
\par\smallskip
\begin{minipage}{0.97\linewidth}\footnotesize Drift-type columns: share of judged Joint conversations whose annotated drift type the judge recovered. $\Delta$: macro-F1 of Joint minus Dual-LLM with the same judge, paired bootstrap CI; all Holm-adjusted $p\le0.021$ (exact McNemar: $p_{\mathrm{Holm}}\le0.024$). Joint vs no OCDD: $\Delta$ between -0.008 and +0.005, all not significant. Joint Phase 1 alone: macro-F1 0.874. Confidence intervals of each framework are shown in Figure~\ref{fig:jointdual}. \$/1k: estimated LLM API cost per 1,000 conversations from the token counts and catalogue prices, including re-requests of outputs truncated by the token limit; local computation is excluded. Claude 5 Sonnet and Nova 2 Lite write long explanations, which, with their prices, raises their cost.\end{minipage}
\end{table*}

The ablation shows where the gain comes from. OCDD detects a drift in \JointDriftShare{} of the conversations flagged by the ensemble (\JointDriftN{} of \JointFlagged{}; held-out: \ExtDriftShare), because flagged conversations are almost all malicious and contain a request for personal information. Removing OCDD therefore changes the final macro-F1 by \DeltaNoOcddMin{} to \DeltaNoOcddMax{} (held-out: \ExtDeltaNoOcddMin{} to \ExtDeltaNoOcddMax; none significant) and saves only \CallsSaved{} of the judge calls. The advantage of the Joint framework comes from its Phase~1 filter.

The recorded predictions show why the judge cannot add accuracy. Under cross-validation the ensemble makes \CompFP{} false alarms and misses \CompFN{} malicious conversations; the judges withdraw \CompFixMin--\CompFixMax{} of the false alarms but also \CompBreakMin--\CompBreakMax{} of the \CompTP{} correct ones, and they never see the missed conversations, although the same LLMs, used as classifiers, recognize \CompFNRightMin--\CompFNRightMax{} of them (Table~\ref{tab:complementarity}) (held-out: \ExtCompFNRightMin--\ExtCompFNRightMax{} of \ExtCompFN). The LLMs thus carry complementary information: an oracle choosing the correct one of the two Phase~1 decisions would reach an accuracy of \CompOracleMin--\CompOracleMax{} instead of \CompEnsAcc. A simple deployable rule does not recover it. Routing the ensemble's non-unanimous votes (\RouteShare{} of the conversations) to an LLM changes macro-F1 by \RouteCVMin{} to \RouteCVMax{} under cross-validation (none significant) and by \RouteExtMin{} to \RouteExtMax{} on the held-out set (significant losses for \RouteExtNSig{} of \RouteExtN{} LLMs); only on the official split, where the ensemble is trained on \NTrain{} conversations, is the change positive for every LLM (\RouteOffMin{} to \RouteOffMax, smallest Holm-adjusted $p=\RouteOffMinP$).

\begin{table*}[tp]
\centering
\caption{Error overlap between the Joint ensemble and each LLM under cross-validation.}
\label{tab:complementarity}
\footnotesize
\begin{tabular}{lccccccc}
\toprule
LLM & \shortstack[c]{LLM\\{}acc.} & \shortstack[c]{FN\\{}rec.} & \shortstack[c]{FP\\{}rec.} & \shortstack[c]{Judge\\{}FP out} & \shortstack[c]{Judge\\{}TP out} & \shortstack[c]{Oracle\\{}acc.} & \shortstack[c]{Held-out\\{}FN rec.} \\
\midrule
Claude 5 Sonnet & 0.828 & 78\% & 33\% & 59\% & 39\% & 0.943 & -- \\
DeepSeek V4 Pro & 0.797 & 74\% & 19\% & -- & -- & 0.930 & -- \\
GPT-6 Luna & 0.815 & 78\% & 22\% & 26\% & 10\% & 0.935 & 83\% \\
GPT-6 Sol & 0.815 & 65\% & 26\% & 15\% & 5\% & 0.930 & -- \\
Gemini 3.1 Pro & 0.807 & 74\% & 30\% & 26\% & 12\% & 0.938 & -- \\
Gemma 4 31B (ASU) & 0.818 & 61\% & 52\% & 11\% & 2\% & 0.945 & 72\% \\
Llama 4 Maverick & 0.805 & 70\% & 33\% & 33\% & 9\% & 0.938 & 78\% \\
Llama 4 Scout (ASU) & 0.772 & 48\% & 52\% & 15\% & 9\% & 0.938 & 56\% \\
Nova 2 Lite & 0.755 & 48\% & 74\% & 52\% & 54\% & 0.953 & -- \\
gpt-oss-120b & 0.748 & 52\% & 67\% & 41\% & 14\% & 0.950 & 44\% \\
\bottomrule
\end{tabular}
\par\smallskip
\begin{minipage}{0.97\linewidth}\footnotesize The ensemble (accuracy 0.875) makes 27 false alarms (FP), 192 correct alarms (TP) and misses 23 malicious conversations (FN). Rec.: share of these conversations that the LLM, used as a zero-shot classifier, labels correctly. Judge FP/TP out: share of the ensemble's false/correct alarms that the LLM as Joint judge does not confirm as a fraudulent attempt. Oracle: accuracy when the correct one of the two Phase~1 decisions is chosen. Held-out: 18 missed conversations; -- marks a model not run in that role.\end{minipage}
\end{table*}

Nor does the detector's output improve the judge (Table~\ref{tab:judgeinputs}). Adding the OCDD drift message to the prompt is no better than adding the fixed fourth message for any of the \NJudgeInput{} judges (\LocWorse{} significantly worse, Holm). Judging only the messages up to the OCDD alarm, an early warning taken on average \EWMsgs{} messages before the end of the dialogue, changes macro-F1 by $+$\EWGainMax{} to $-$\EWLossMax{} relative to the whole dialogue, and a prefix ending at the fixed fourth message performs alike, because in SEConvo the request almost always comes early.

\begin{table*}[tp]
\centering
\caption{Judge-input experiment (5-fold CV, Joint framework): macro-F1 of the final \emph{fraudulent attempt} decision when the judge receives the whole dialogue, the dialogue plus the drift location (from OCDD or the fixed fourth message), or only the messages up to the alarm.}
\label{tab:judgeinputs}
\footnotesize
\begin{tabular}{lccccc}
\toprule
Judge LLM & \shortstack[c]{Whole\\{}dialogue} & \shortstack[c]{+ OCDD\\{}location} & \shortstack[c]{+ fixed\\{}location} & \shortstack[c]{Prefix to\\{}OCDD alarm} & \shortstack[c]{Prefix to\\{}message 4} \\
\midrule
Gemma 4 31B (ASU) & 0.872 & 0.872 & 0.877 & 0.870 & 0.872 \\
Llama 4 Maverick & 0.852 & 0.850 & 0.865 & 0.825 & 0.822 \\
GPT-6 Luna & 0.842 & 0.825 & 0.825 & 0.852 & 0.862 \\
gpt-oss-120b & 0.837 & 0.845 & 0.860 & 0.805 & 0.802 \\
Llama 4 Scout (ASU) & 0.842 & 0.870 & 0.869 & 0.774 & 0.767 \\
\bottomrule
\end{tabular}
\par\smallskip
\begin{minipage}{0.97\linewidth}\footnotesize Fixed variants are controls that use no detector output. Paired bootstrap of OCDD minus fixed: no judge improves (Holm-adjusted).\end{minipage}
\end{table*}

\section{Discussion}\label{sec:discussion}

The ensemble generalizes better than zero-shot LLMs because the two read the task differently: LLMs apply a broad notion of ``suspicious'' and flag outreach whose requests are proportionate, whereas a supervised model learns the boundary that human labels draw, and few-shot prompting narrows the gap. Because all SEConvo conversations come from one generator, lexical regularities may favour TF-IDF models, and the held-out set does not test other generators or real users.

Drift detection inside short dialogues depends above all on the reference against which a message is judged. The fraudulent shift typically occurs in the initiator's second message, so a one-class model fitted on the dialogue's own history has one or two messages to learn from and fires at the first message it can score; the comparison with a fixed fourth-message rule exposes this behaviour. A reference of normal messages pooled across training dialogues, represented by sentence embeddings, avoids the problem and separates drifted from stable dialogues. Its weaker result on the official split is plausibly related to the seven dialogues without a drift in that training part, too few to estimate the false-drift rate during configuration selection, although the smaller training sample as a whole may also contribute.

The results also separate two roles of OCDD. As a detector, measured against the first PII request as onset proxy, it is significantly better than the fixed-position rule and the window reference and the only method that follows a late request. As a pipeline component it adds nothing measurable in SEConvo, because nearly every flagged conversation contains a request and the judge gains neither from the drift location nor from deciding at alarm time. Both observations stem from a corpus in which requests come early and at regular positions; the detector's downstream value remains to be shown where the onset varies and flagged conversations often lack a shift, as in longer dialogues or spam-heavy streams.

The judges' tendency to read legitimate requests as fraud is plausibly encouraged by the judge prompt, which presents the input as a \emph{detected concept drift} and asks for a binary verdict. Our judge-input experiment shows that adding the drift location alone does not change this; examples of legitimate requests, or calibrated confidence with an option to abstain~\cite{senol2026abstain,senol2026epistemic}, are more promising, and behavioural measures of reasoning quality~\cite{senol2026reasoning} could guide judge selection. The error analysis also locates the unused potential of the LLMs: they recognize many of the attacks the ensemble misses, which a judge of flagged conversations never sees, but a vote-based routing rule exploits this only when the ensemble is trained on few labels, so a calibrated router for the conversations the ensemble clears is a more promising use of the LLM than a second check of its alarms. In practice, the Joint framework confines LLM use to flagged conversations, and the best cost--quality balance was obtained with university-hosted open-weight judges, which also keeps sensitive dialogues away from third-party APIs. Provider content filters, as observed for Nova~2~Lite, are an operational risk for hosted judges.

This study has limitations. We use a single synthetic English corpus; the held-out set comes from the same generator and its labels encode simulation intent, so it tests unseen conversations rather than new sources. The first PII request is an operational proxy for the onset of a shift rather than a direct annotation of concept drift, and the evaluation is retrospective rather than a live deployment. With the \NTrain{} official training conversations the ensemble relies on a representation fitted on unlabelled test texts (macro-F1 \JointMF{} transductively against \JointIndMF{} inductively), which requires a batch of unlabelled conversations, although the advantage shrinks as labels grow and an inductive SVM trained on the \NConv{} annotated conversations reaches \ExtIndSVM{} on the held-out set. Finally, we use one judge prompt and one decoding setting, hosted LLMs change over time, and DeepSeek was evaluated as a classifier but not as a judge.

\section{Conclusions}\label{sec:conclusion}

We evaluated a two-phase framework for detecting social-engineering conversations: a majority-vote ensemble filter followed by within-dialogue OCDD and an LLM judge of the drift type, compared with a Dual-LLM alternative across \NLLM{} current LLMs, annotated drift onsets, cross-validation and a held-out test set from the same corpus. The ensemble is on par with the best LLM classifiers on the annotated data and clearly better on unseen conversations (H1). OCDD with a normal-message reference and sentence embeddings localizes the shift in \OcddLocal{} of drifted dialogues and detects drift well above the reference value of 0.5 (balanced score \OcddBal), whereas a dialogue's own recent messages are no better than a fixed-position rule; LLM judges, however, over-label legitimate requests as fraudulent (H2, partly supported). The Joint framework is significantly more accurate than Dual-LLM for every judge under cross-validation and on held-out data, at a fraction of the LLM cost (H3). This advantage comes from the Phase~1 filter, and the judge cannot recover the attacks the filter misses, although the LLMs recognize many of them as classifiers; in SEConvo, where requests come early and at regular positions, neither removing OCDD nor passing its output to the judge changes the final decision. These results favour a lightweight supervised filter with selective LLM judgment over LLM-only pipelines. Future work should test the drift detector where the onset varies, in longer or spam-heavier streams, calibrate judges on legitimate requests, and validate the framework on human conversations.

\backmatter

\bmhead{Acknowledgements}
The authors gratefully acknowledge the support of The Scientific and Technological Research Council of T\"urkiye (T\"UB\.ITAK). LLM access was provided through the Arizona State University CreateAI platform.

\section*{Statements and Declarations}

\bmhead{Funding}
This research was supported by The Scientific and Technological Research Council of T\"urkiye (T\"UB\.ITAK) under the 2219 International Postdoctoral Research Fellowship Program.

\bmhead{Competing interests}
The authors have no competing interests to declare that are relevant to the content of this article.

\bmhead{Ethics approval and consent to participate}
Not applicable. The study uses only a publicly available corpus of synthetic conversations and involves no human participants.

\bmhead{Data availability}
SEConvo is available from Zenodo (\href{https://doi.org/10.5281/zenodo.12170260}{https://doi.org/10.5281/zenodo.12170260}, CC BY 4.0)~\cite{seconvo2024data}. The per-conversation outputs and summary tables of the reported experiments are available in the code repository below.

\bmhead{Code availability}
The implementation, configuration and prompts, and the scripts that regenerate every table and figure of this paper are publicly available at \url{https://github.com/senolali/jointframework} under the MIT license.

\bmhead{Author contributions}
Conceptualization: A.\c{S}. and H.L.; Methodology: A.\c{S}. and G.A.; Software: A.\c{S}.; Validation: A.\c{S}., G.A. and H.L.; Formal analysis: A.\c{S}.; Investigation: A.\c{S}.; Resources: H.L.; Visualization: A.\c{S}.; Writing -- original draft: A.\c{S}.; Writing -- review and editing: G.A. and H.L.; Supervision: H.L.; Project administration: A.\c{S}.; Funding acquisition: A.\c{S}. All authors read and approved the final manuscript.

\edef\maintables{\the\value{table}}
\begin{appendices}

\section{Supplementary tables}\label{app:tables}
\renewcommand{\thetable}{\arabic{table}}\setcounter{table}{\maintables}

Table~\ref{tab:notation} explains the abbreviations and symbols used in the paper. Table~\ref{tab:config} gives the complete experimental configuration, Table~\ref{tab:hyper} the search grids and selected hyper-parameters of the Phase~1 candidates, and Table~\ref{tab:phase2official} the Phase~2 results on the official split.

\begin{table*}[tp]
\centering
\caption{Abbreviations and symbols.}
\label{tab:notation}
\footnotesize
\begin{tabularx}{\linewidth}{@{}lX@{}}
\toprule
\textbf{Term} & \textbf{Meaning} \\
\midrule
\multicolumn{2}{@{}l}{\emph{Abbreviations}} \\
CI & Confidence interval (95\%, bootstrap) \\
CV & Cross-validation \\
DBSCAN & Density-based spatial clustering of applications with noise \\
GMM & Gaussian mixture model \\
kNN & $k$-nearest neighbours \\
LLM & Large language model \\
NB & Naive Bayes \\
OCDD & One-Class Drift Detector \\
OC-SVM & One-class support vector machine \\
PCA / SVD & Principal component analysis (sentence embeddings) / truncated singular value decomposition (TF-IDF vectors) \\
PII & Personally identifiable information (e.g., date of birth, address, e-mail) \\
RBF & Radial basis function (kernel) \\
RF & Random Forest \\
SVM & Support vector machine \\
TF-IDF & Term frequency--inverse document frequency \\
\midrule
\multicolumn{2}{@{}l}{\emph{Symbols}} \\
$C=(m_1,\dots,m_T)$ & A conversation of $T$ messages $m_t$ \\
$x_t$ & Vector representation of message $m_t$ \\
$d$ & Dimension of the message representation after SVD or PCA \\
$t_0$ & Message at which a dialogue shifts, approximated by the first PII request \\
$t^\ast$ & Message at which OCDD declares a drift, the estimate of $t_0$ \\
$\nu$ & One-class SVM parameter: upper bound on the share of training messages treated as outliers \\
$q$ & Quantile of held-out normal-message scores used as the outlier threshold \\
$o_t$ & Outlier flag of message $m_t$ (1 if its score exceeds the threshold) \\
$\ell$ & Number of most recent scored messages over which outliers are counted \\
$\rho$ & Share of outliers among the last $\ell$ messages at which a drift is declared \\
$W$ & Number of recent messages forming the reference in the dialogue-window variant \\
$\Delta$ & Difference in macro-F1 between two systems evaluated on the same conversations \\
\midrule
\multicolumn{2}{@{}l}{\emph{Evaluation terms}} \\
First PII request & First message of a conversation that asks for PII; used as the drift onset \\
Localization rate & Share of drifted dialogues in which the drift is flagged at the onset or up to two messages later \\
False-drift rate & Share of dialogues without a drift in which a drift is flagged \\
Balanced score & Mean of the localization rate and one minus the false-drift rate; 0.5 is a nominal reference value (a detector that never fires scores 0.5, one that always fires at most 0.5) \\
Macro-F1 & Unweighted mean of the F1 scores of the malicious and benign classes \\
$p_{\mathrm{Holm}}$ & $p$-value after Holm adjustment for multiple comparisons \\
\bottomrule
\end{tabularx}
\end{table*}

\begin{table*}[tp]
\centering
\caption{Experimental configuration (generated from the run's configuration file).}
\label{tab:config}
\footnotesize
\begin{tabularx}{\linewidth}{@{}p{3.1cm}X@{}}
\toprule
\textbf{Component} & \textbf{Setting} \\
\midrule
Text representation & TF-IDF on word 1--2-grams, sublinear tf, English stop words, min.\ document frequency 2, at most 20,000 terms; truncated SVD to 10 components and $\ell_2$ normalization. Fitted inductively (training texts) and transductively (all texts, no labels). \\
OCDD message spaces & TF-IDF + truncated SVD ($d\in\{5, 10, 20\}$) and PCA-reduced te3-large embeddings ($d\in\{10, 20, 50\}$) of individual messages, fitted without labels. \\
OCDD search grid & design $\in$ \{normal-message reference, shift from the opening message\}; outlier threshold at the $q$-quantile of held-out normal-message scores, $q\in\{0.5, 0.7, 0.8, 0.9, 0.95\}$; $(\ell,\rho)\in\{(1,1.0), (2,0.5), (2,1.0)\}$; one-class SVM $\nu\in\{0.1, 0.5\}$, RBF kernel, $\gamma$ = scale, standardized inputs; nested 5-fold cross-fitting within the training dialogues. \\
OCDD selected & normal-message reference (initiator messages before the first PII request in the training dialogues); embeddings, $d=10$; $\nu=0.5$; $q=0.9$; $\ell=2$, $\rho=0.5$ (selected on all 400 annotated dialogues; identical design in all CV folds); monitored: initiator messages after the opening one; localization tolerance 2 messages. \\
LLM decoding & temperature 0.0 where accepted (GPT-6 models reject the parameter); maximum output tokens 600 (Claude, Llama, Gemma), 1,000 (Nova), 3,000 (GPT-6, Gemini, gpt-oss), 4,000 (DeepSeek); outputs cut by the limit are re-requested once with a 4$\times$ budget (cap 32,000). \\
LLM classification & JSON output (label, confidence, evidence turns, rationale); unparseable outputs counted as benign; few-shot: 4 class-balanced training examples; 1 run per input. \\
LLM judge & judge prompt (Figure~\ref{fig:prompt}), dialogue as JSON; verdict parsed from `fraudulent' / `genuine'. \\
Software & Python 3.13.11, scikit-learn 1.9.1, impkmeans, XGBoost 3.4.1; LLMs via the ASU CreateAI OpenAI-compatible API (DeepSeek via its own API). \\
\bottomrule
\end{tabularx}
\end{table*}
\begin{table*}[tp]
\centering
\caption{Phase 1 model configurations. Hyper-parameters are selected by 5-fold stratified inner cross-validation on the training conversations (criterion: macro-F1); the last column gives the values selected on the official split with the transductive TF-IDF/SVD representation.}
\label{tab:hyper}
\scriptsize
\begin{tabular}{p{2.0cm}p{3.4cm}p{4.6cm}p{3.4cm}}
\toprule
Model & Fixed settings & Inner-CV search grid & Selected (official split) \\
\midrule
kNN & standardized; Euclidean & n\_neighbors $\in$ \{3, 5, 7, 9\}; weights $\in$ \{uniform, distance\} & n\_neighbors=3, weights=distance \\
SVM & standardized; class-balanced; probability outputs & kernel $\in$ \{rbf, linear\}; C $\in$ \{0.1, 1.0, 10.0\} & C=1.0, kernel=rbf \\
Random Forest & 300 trees; class-balanced & max\_depth $\in$ \{None, 6\}; max\_features $\in$ \{sqrt, 0.3\} & max\_depth=None, max\_features=sqrt \\
Naive Bayes & Complement NB on TF-IDF counts & alpha $\in$ \{0.1, 0.5, 1.0\} & alpha=0.1 \\
XGBoost & 300 trees; learning rate 0.05; subsample 0.9 & max\_depth $\in$ \{2, 4\}; colsample\_bytree $\in$ \{0.5, 1.0\} & colsample\_bytree=0.5, max\_depth=2 \\
k-means & 10 initializations & n\_clusters $\in$ \{2\} & n\_clusters=2 \\
DBSCAN & $\varepsilon$ = median $k$-distance & min\_samples $\in$ \{3, 5\} & min\_samples=3 \\
GMM & diagonal covariance & n\_components $\in$ \{2\} & n\_components=2 \\
ImpKmeans & official package; KDE in the 10-dimensional SVD space; 1 k-means iteration & n\_clusters $\in$ \{2\}; r $\in$ \{0.3, 0.6, 1.0\}; bandwidth $\in$ \{0.05, 0.2\} & bandwidth=0.2, n\_clusters=2, r=0.6 \\
\bottomrule
\end{tabular}
\end{table*}
\begin{table*}[tp]
\centering
\caption{Phase 2 on the official test split (360 dialogues), with the OCDD configuration selected on the 40 training dialogues.}
\label{tab:phase2official}
\footnotesize
\begin{tabular}{lcccc}
\toprule
Judge LLM & Joint [95\% CI] & No OCDD & Dual-LLM & $p_{\mathrm{Holm}}$ \\
\midrule
GPT-6 Sol & 0.830 [0.789, 0.868] & 0.841 & 0.814 & 1.000 \\
Llama 4 Maverick & 0.822 [0.781, 0.861] & 0.833 & 0.807 & 1.000 \\
Gemma 4 31B (ASU) & 0.816 [0.773, 0.855] & 0.827 & 0.821 & 1.000 \\
GPT-6 Luna & 0.800 [0.756, 0.844] & 0.811 & 0.795 & 1.000 \\
gpt-oss-120b & 0.800 [0.758, 0.839] & 0.811 & 0.752 & 0.480 \\
Gemini 3.1 Pro & 0.792 [0.749, 0.833] & 0.803 & 0.771 & 1.000 \\
Llama 4 Scout (ASU) & 0.775 [0.730, 0.817] & 0.783 & 0.758 & 1.000 \\
Claude 5 Sonnet & 0.679 [0.631, 0.728] & 0.685 & 0.689 & 1.000 \\
Nova 2 Lite & 0.614 [0.559, 0.666] & 0.624 & 0.585 & 1.000 \\
\bottomrule
\end{tabular}
\par\smallskip
\begin{minipage}{0.97\linewidth}\footnotesize $p_{\mathrm{Holm}}$: exact McNemar test Joint vs Dual-LLM (same judge), Holm-corrected.\end{minipage}
\end{table*}

\end{appendices}
\end{document}